%% file: sf_workshop.tex
\documentclass{article}

\usepackage[utf8]{inputenc}

\usepackage{times}
\usepackage{graphicx} 
\usepackage{wrapfig}
\usepackage{subfig}
\usepackage{caption}
\usepackage{makeidx}
\usepackage{multirow}
\usepackage{chngpage}

\usepackage{todonotes}

\usepackage{natbib}

\usepackage{algorithm}
\usepackage{algorithmic}

\usepackage{hyperref}

\usepackage[accepted]{icml2017}

\usepackage{amsmath}
\usepackage{amsfonts}
\usepackage{amssymb}
\usepackage{proof}
\usepackage{amsthm}
\usepackage{algorithm,algorithmic}

\newcommand{\norm}[1]{\left|\left|#1\right|\right|}

\icmltitlerunning{Advantages and Limitations of using Successor Features for Transfer in Reinforcement Learning}

\begin{document} 

\twocolumn[
\icmltitle{Advantages and Limitations of using Successor Features \\ for Transfer in Reinforcement Learning}



\icmlsetsymbol{equal}{*}

\begin{icmlauthorlist}
\icmlauthor{Lucas Lehnert}{ed}
\icmlauthor{Stefanie Tellex}{ed}
\icmlauthor{Michael L.\ Littman}{ed}
\end{icmlauthorlist}

\icmlaffiliation{ed}{Brown University, Providence, Rhode Island, USA}

\icmlcorrespondingauthor{Lucas Lehnert}{lucas\_lehnert@brown.edu}

\icmlkeywords{Reinforcement Learning, Successor Features, Transfer}

\vskip 0.3in
]



\printAffiliationsAndNotice{}  

\begin{abstract}
One question central to Reinforcement Learning is how to learn a feature representation that supports algorithm scaling and re-use of learned information from different tasks.
Successor Features approach this problem by learning a feature representation that satisfies a temporal constraint.
We present an implementation of an approach that decouples the feature representation from the reward function, making it suitable for transferring knowledge between domains. We then assess the advantages and limitations of using Successor Features for transfer. 
\end{abstract}

\section{Introduction}

Reinforcement Learning (RL)~\cite{kaelbling1996reinforcement,Sutton98} studies the problem of computing an optimal control strategy using one-step interactions sampled from an environment.
For each selected action, the environment also provides a reward, a single scalar number.
The goal is to compute a control strategy, also called a policy, that maximizes the cumulative reward received while interacting with the environment.
One challenge in this setting is transferring knowledge about one environment to another when only the reward specification changes, but the remaining specification of the environment stays fixed.
In this paper, we consider the approach presented by~\citet{barreto2016successor}, which uses Successor Features (SF) to compute a representation of the environment that can be transferred across different reward functions.
We present an implementation of this method and show that while learning a SF representation has significant benefits for transfer, it has also some fundamental limitations.

\section{Background}

We consider a \emph{Markov Decision Process (MDP)} $M = \langle \mathcal{S}, \mathcal{A}, p, r, \gamma \rangle$ with a finite state space $\mathcal{S}$ and a finite action space $\mathcal{A}$. 
The transition function $p$ specifies with $p(s,a,s')$ the probability of transitioning from a state $s$ to a state $s'$ when selecting an action $a$.
For every such transition, the reward is specified by the reward function $r: \mathcal{S} \times \mathcal{A} \to \mathbb{R}$.
Further, we assume a discount factor $\gamma \in [0,1)$ that weights the tradeoffs between immediate and long term rewards.

Let $\pi$ be a policy that specifies the distribution with which actions are selected, conditioned on the state space $\mathcal{S}$.
The Q-function of this policy is defined as
\begin{equation}
Q^\pi(s,a) = \mathbb{E}_\pi \left[ \sum_{t=1}^\infty \gamma^{t-1} r_t \middle| s_1 = s, a_1 = a \right], \label{eq:q-fn}
\end{equation}
where the expectation $\mathbb{E}_\pi$ is over all possible infinite length trajectories in $M$ and $r_t$ the reward at time step $t$. 

Several algorithms have been developed to estimate a Q-function, however, one important question is how to represent a current Q-function estimate.
For example, suppose the state space of an MDP $M$ consists of $n$ states and $m$ actions, then an estimate of the Q-function can be stored in a vector $\pmb{\theta}$ of dimension $mn$:
\begin{equation}
\pmb{\theta} = \left[ Q(s_1,a_1), \cdots ,Q(s_n,a_m) \right]^\top \label{eq:q-tab}
\end{equation} 
To compute the Q-value for a state-action pair $(s,a)$, a basis function 
\begin{equation}
\psi: (s,a) \mapsto \pmb{\psi}_{s,a}
\end{equation}
can be used, where $\pmb{\psi}_{s,a}$ is a one-hot bit vector of dimension $mn$.
Basis functions can also be generalized to have different forms to further improve scalability of different learning algorithms~\cite{sutton1996generalization,konidaris2008fourier}.

\section{Learning Successor Features for Transfer}

\citet{dayan1993successor} presented Successor Features (SFs), a particular type of basis function that represents a state as a feature vector $\pmb{\psi}^\pi_{s,a}$ such that under a given policy the feature representation $\pmb{\psi}^\pi_{s,a}$ is similar to the feature representation of its successor states.
The idea originates from the Bellman fixed-point equation,
\begin{equation}
Q^\pi(s,a) = r(s,a) + \gamma \mathbb{E}_{s',a'}  \left[ Q^\pi(s',a') \right], \label{eq:Q-bellman-fix}
\end{equation}
where $s'$ is the sampled next state and $a'$ is the sampled next action at state $s'$.
If the Q-function is approximated linearly, then 
\begin{equation}
(\pmb{\psi}^\pi_{s,a})^\top \pmb{\theta} \approx r(s,a) + \gamma \mathbb{E}_{s',a'}  \left[ (\pmb{\psi}^\pi_{s',a'})^\top \pmb{\theta} \right]. \label{eq:q-fix-approx}
\end{equation}
Note that, depending on the choice of basis function, \eqref{eq:q-fix-approx} may not hold exactly because we only estimate a linear approximation of the true Q-function.
The objective of finding a good SF representation is to find a basis function $\psi$ such that~\eqref{eq:q-fix-approx} holds as exactly as possible.

\citet{barreto2016successor} re-visited this approach in the context of transferring a feature representation within a set of MDPs where only the reward function varies.
While various different approaches were presented to this problem (see \citet{stone2009transfersurvey} for a survey),~\citeauthor{barreto2016successor} approach this transfer problem by learning a feature representation that is descriptive of the entire set of MDPs and can be used for transfer across different reward functions.

Intuitively, the Q-function combines information about the reward function itself, as well as the temporal ordering of the received rewards.
This temporal ordering is induced by the current policy $\pi$ and the transition dynamics of the MDP that determine which trajectories are generated.

For transfer, \citeauthor{barreto2016successor} present an approach that isolates the reward function from the Q-function.
They define a basis function $\phi : (s,a) \mapsto \pmb{\phi}_{s,a}$ to parametrize the reward function with
\begin{equation}
r(s,a) = \pmb{\phi}_{s,a}^\top \pmb{w}. \label{eq:reward-model}
\end{equation}
Since~\eqref{eq:reward-model} is stated as a strict equality, the assumption is made that $\phi$ is not too restrictive and the reward function $r$ can be represented exactly.
Using this assumption, \citeauthor{barreto2016successor} rewrite the Q-function as
\begin{align}
Q^\pi(s,a) &= \mathbb{E}_\pi \left[ \sum_{t=1}^\infty \gamma^{t-1} r_t  \middle| s_0=s, a_0=a \right] \nonumber \\
&= \mathbb{E}_\pi \left[ \sum_{t=1}^\infty \gamma^{t-1}  \pmb{\phi}_t^\top \pmb{w}  \middle| s_0=s, a_0=a \right] \nonumber \\
&= \underbrace{\mathbb{E}_\pi \left[ \sum_{t=1}^\infty \gamma^{t-1} \pmb{\phi}_t \middle| s_0=s, a_0=a \right]^\top}_{\overset{\text{def.}}{=} (\pmb{\psi}^\pi_{s,a})^\top} \pmb{w}, \label{eq:Q-fn-split}
\end{align}
where $\pmb{\phi}_t$ is the reward feature at time step $t$ for a trajectory started at $(s,a)$. 
Suppose $\phi$ is a basis function that tabulates the state-action space, i.e. $\pmb{\phi}_{s,a}$ is a one-hot bitvector of dimension $| \mathcal{S} \times \mathcal{A} |$.
In this case, the weight vector $\pmb{w}$ can be thought of as the full reward model written out as a vector.
This means~\eqref{eq:Q-fn-split} can be interpreted as a separation of the Q-function into a (linear) factor $\pmb{w}$ describing rewards only and a (linear) factor describing the ordering with which rewards are observed.
Hence,~\citeauthor{barreto2016successor} propose to learn a \emph{Successor Feature} $\psi : (s,a) \mapsto \pmb{\psi}_{s,a}$ satisfying 
\begin{align}
\pmb{\psi}^\pi_{s,a} &= \mathbb{E}_\pi \left[ \sum_{t=0}^\infty \gamma^t \pmb{\phi}_t \middle| s_0=s, a_0=a \right]\nonumber \\
&= \pmb{\phi}_{s,a} + \gamma \mathbb{E}_{s',a'} \left[ \pmb{\psi}^\pi_{s',a'} \right].  \label{eq:SF-def}
\end{align}
In addition, they also present policy improvement theorems similar to the usual dynamic programming improvement theorems~\cite{Sutton98}.

\subsection{Algorithm Derivation}

Similar to Fitted Q-iteration~\cite{antos2006fittedQ}, DQN~\cite{mnih2015human}, and the method outlined by~\citet{zhang2016deep}, we derive a learning algorithm that fits a reward model and SF model by simultaneously minimizing two loss functions.
The reward model is fitted by minimizing the reward loss
\begin{equation}
\mathcal{L}_R(\pmb{\phi},\pmb{w}) = \mathbb{E}_{s,a} \left[ \norm{\pmb{\phi}_{s,a}^\top \pmb{w} - r_{s,a}}^2 \right], \label{eq:reward-loss}
\end{equation}
where the expectation $\mathbb{E}_{s,a}$ is with respect to some visitation distribution over the state-action space $\mathcal{S} \times \mathcal{A}$, and where the scalar $r_{s,a}$ is the reward received for a particular transition.

The SF $\psi$ is learned by first estimating a target 
\begin{equation}
\pmb{y}_{s,a,s'} = \begin{cases} \pmb{\phi}_{s,a} &\text{if $s'$ is terminal} \\ \pmb{\phi}_{s,a} + \gamma \mathbb{E}_{a'} \left[ \pmb{\psi'}_{s',a'}^\pi \right] &\text{otherwise} \end{cases}
\end{equation}
for every collected transition $(s,a,s')$.
For computing this target, the SF estimate $\pmb{\psi'}$ of the previous update iteration is used.
Unlike~\citeauthor{mnih2015human}'s Deep Q-learning, the target $\pmb{y}_{s,a,s'}$ is a vector and not a single scalar variable.
For learning a SF representation, the loss objective
\begin{equation}
\mathcal{L}_{SF}(\pmb{\psi}) = \mathbb{E}_{s,a,s'} \left[ \norm{ \pmb{\psi}_{s,a} - \pmb{y}_{s,a,s'} }^2 \right] \label{eq:sf-loss}
\end{equation}
is used.
The gradient of~\eqref{eq:sf-loss} with respect to the parameters $\pmb{\theta}$ is 
\begin{equation}
\nabla_{\pmb{\theta}} \mathcal{L}_{SF}(\pmb{\psi}) = 2 \mathbb{E}_{s,a,s'} \left[ ( \pmb{\psi}_{s,a} - \pmb{y}_{s,a,s'} ) \nabla_{\pmb{\theta}} \pmb{\psi}_{s,a} \right], \label{eq:sf-loss-grad}
\end{equation}
which is similar to the gradient used by Deep Q-learning with the distinction that~\eqref{eq:sf-loss-grad} is a matrix rather than a vector, and~\eqref{eq:sf-loss} is defined on the SF $\psi^\pi$, rather than Q-values.

Algorithm~\ref{alg:fsf} outlines the implemented SF learning method.
Learning is stabilized by sampling a batch of transitions and using the entire batch to make a gradient descent update.

\begin{algorithm}     
\caption{Fitted SF Learning}
\label{alg:fsf}     
\begin{algorithmic}
\STATE Initialize $\psi$, $\phi$, and $\pmb{w}$.
\LOOP 
\STATE Collect transitions $\tau = \{ (s_t,a_t,r_t,s_{t+1}) \}_{t=T}^{T+N}$ using the Q-function estimate $Q(s,a) = (\pmb{\psi}_{s,a}^\pi)^\top \pmb{w}$
\STATE Using $\tau$ perform gradient update on $\mathcal{L}_R(\pmb{\phi},\pmb{w})$ and $\mathcal{L}_{SF}(\pmb{\psi})$
\ENDLOOP
\end{algorithmic}
\end{algorithm}

\section{Experiments: Grid World}

Algorithm~\ref{alg:fsf} is first evaluated on a $10 \times 10$ grid world navigation task with four actions: up, down, left, or right.
Transitions are stochastic and with a 5\% probability the agent moves sideways.
Rewards are set to 1 for entering the goal cell (terminal state) in the top right corner, and otherwise a zero reward is given.
Every episode is started in the bottom right corner and the discount factor is set to $\gamma = 0.9$.
Actions are selected using an $\varepsilon$-greedy policy with respect to the current Q-value estimates: with probability $\varepsilon=0.3$ actions are selected uniformly at random and with probability $1 - \varepsilon$ the action with the highest Q-value estimate is used.

We compare our Fitted SF implementation against a Fitted Q-iteration implementation.
To ensure a fair comparison, Fitted Q-iteration is identical to Fitted SF except that Fitted Q-iteration minimizes the loss objective 
\begin{equation}
\mathcal{L}_Q(Q_{\pmb{\theta}}) = \mathbb{E}_{s,a,s'} \left[ \norm{ Q_{\pmb{\theta}}(s,a) - y_{s,a,s'} }^2 \right],
\end{equation}
where the target is set to 
\begin{equation}
y_{s,a,s'} = \begin{cases} r_{s,a} &\text{if $s'$ is terminal} \\ r_{s,a} + \gamma V_{\pmb{\theta}'}(s') & \text{otherwise.}\end{cases}
\end{equation}
The value estimate $V_{\pmb{\theta}'}(s') = \max_{a'} Q_{\pmb{\theta}'}(s',a')$ and $Q_{\pmb{\theta}'}$ is the Q-function estimate of the previous iteration.

In all experiments, the Q-function in Fitted Q-iteration uses a basis function tabulating the state action space and the weight vector $\pmb{\theta}$ is learned as described in~\eqref{eq:q-tab}.
Further, the basis function $\phi$ used for estimating the reward model~\eqref{eq:reward-model} also tabulates the state-action space; that is, the reward model can always exactly represent the true reward function.
The SF representation is learned as a linear transform on the tabular basis function $\phi$:
\begin{equation}
\pmb{\psi}_{s,a} = \pmb{\Psi} \pmb{\phi}_{s,a}.
\end{equation}
Because all basis functions are chosen to be tabular, and SFs are linear in a tabular one-hot basis function, both algorithms are not constrained in their representation and can always capture the true value function, reward model, and successor features.

\subsection{Single Task Learning}

\begin{figure}
\centering

\includegraphics[width=.48\textwidth]{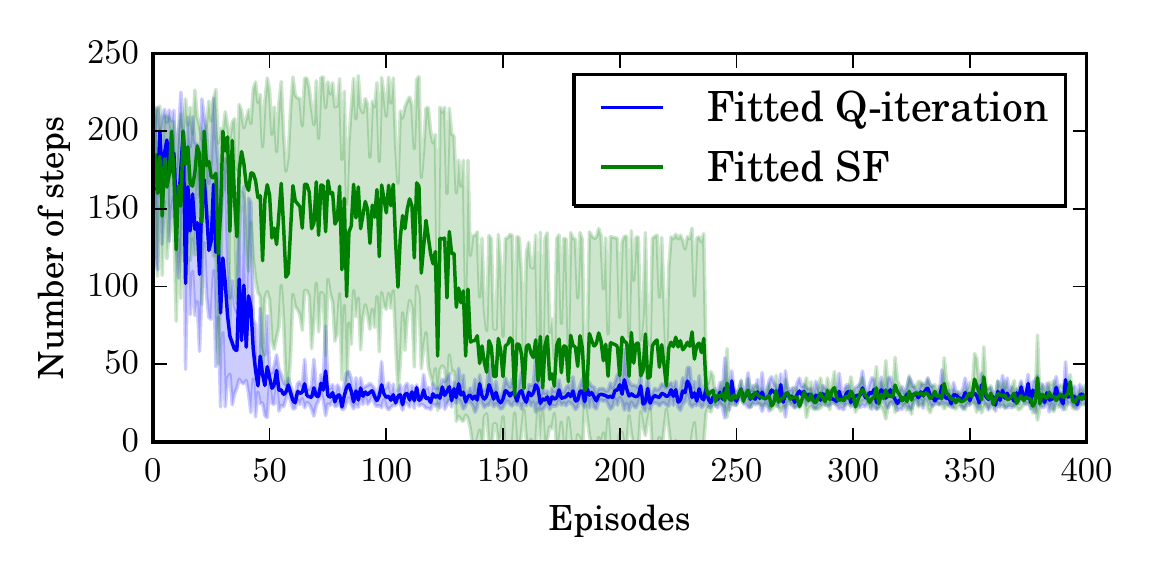}
\vspace{-.9cm}
\caption{Episode length for the best Fitted Q-iteration run and Fitted SF run. All experiments were repeated 20 times and the average episode length plus standard deviation is plotted. The shorter the episode, the sooner the agent can reach the +1 reward state---a shorter episode is better.}
\label{fig:single-run-comp}
\end{figure}

\begin{figure}[h!]
\centering
\vspace{-0.75cm}
\subfloat[Loss Objective $\mathcal{L}_{SF}$~\eqref{eq:sf-loss}]{\label{fig:single-run-lsf}\includegraphics[width=0.24\textwidth]{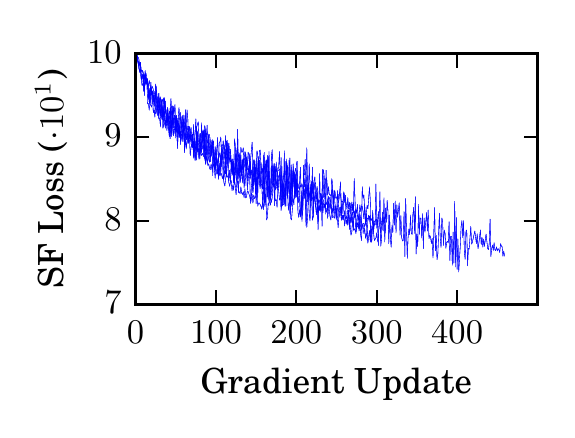}}
\subfloat[Loss Objective $\mathcal{L}_{R}$~\eqref{eq:reward-loss}]{\label{fig:single-run-r}\includegraphics[width=0.24\textwidth]{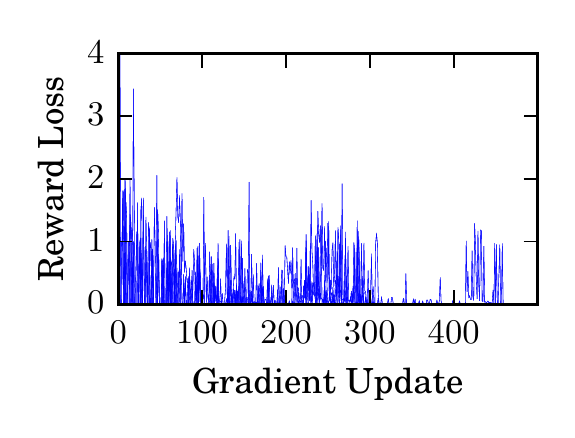}}
\caption{Evolution of the loss objectives.
Fitted SF minimizes using the Adagrad gradient descent optimizer implemented in Tensorflow~\cite{tensorflow2015-whitepaper}. 
A learning rate of $0.01$ performed best for the loss objective $\mathcal{L}_{SF}$ and a learning rate of $0.1$ performed best for the loss objective $\mathcal{L}_R$. The fitted Q-iteration implementation performed best with a learning rate of $0.01$.
Otherwise Tensorflow's default parameters were used.}
\label{fig:single-run-loss}
\end{figure}

Figure~\ref{fig:single-run-comp} compares the performance of the Fitted SF algorithm against Fitted Q-iteration.
Both algorithms converge to a good solution and can perform the navigation task in few steps at the end of training\footnote{Note that the control policy was constrained to be only $\varepsilon$-greedy with $\varepsilon=0.3$.}.
The Fitted SF algorithm converges slower, which can be explained by the fact that it has to learn a full reward model before it can form good Q-value estimates.
Figure~\ref{fig:single-run-loss} shows that the Fitted SF algorithm robustly minimizes both its loss objectives.

\begin{figure*}
\centering
\subfloat[Comparison of Fitted SF learning with Fitted Q-iteration. 
Fitted Q-iteration used a learning rate of $0.1$, Fitted SF learning used a learning rate of $0.0001$ for the SF and a learning rate of $0.1$ for the reward model.]{\label{fig:tiny-change-none}\label{fig:tiny-change-comp-a}\includegraphics[width=.48\textwidth]{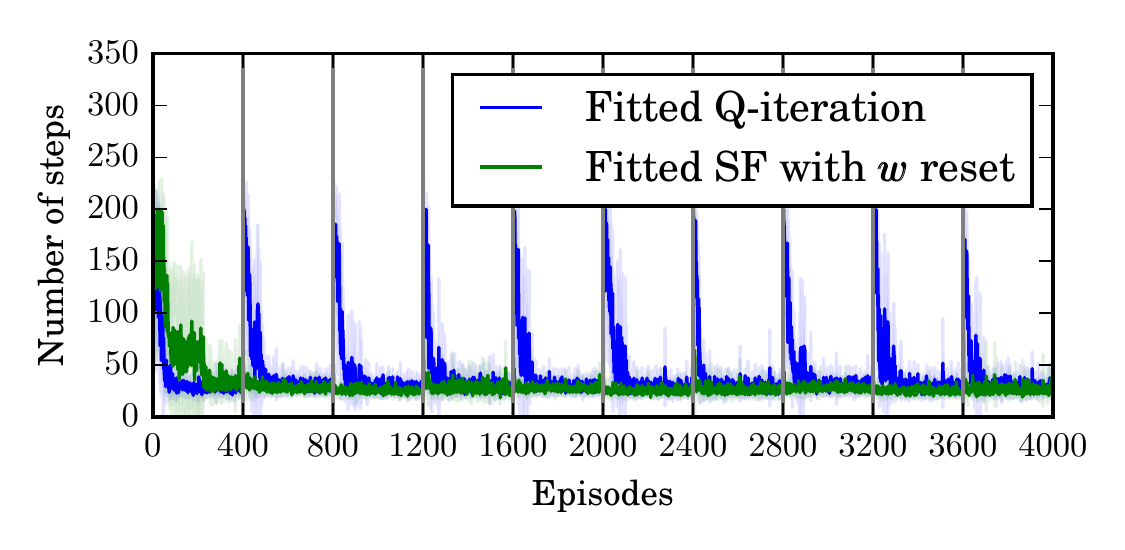}}~
\subfloat[Comparison of different weight resetting strategies for the SF algorithm. The green curve is the same as in Figure~\ref{fig:tiny-change-none}. The blue curve shows the episode length when all weights are reinitialized between training rounds, the green curve keeps the matrix $\pmb{\Psi}$ between reward function changes. 
The blue curve used a learning rate of 0.001 for the SF and 0.01 for the reward model.
]{\label{fig:tiny-change-comp-b}\includegraphics[width=.48\textwidth]{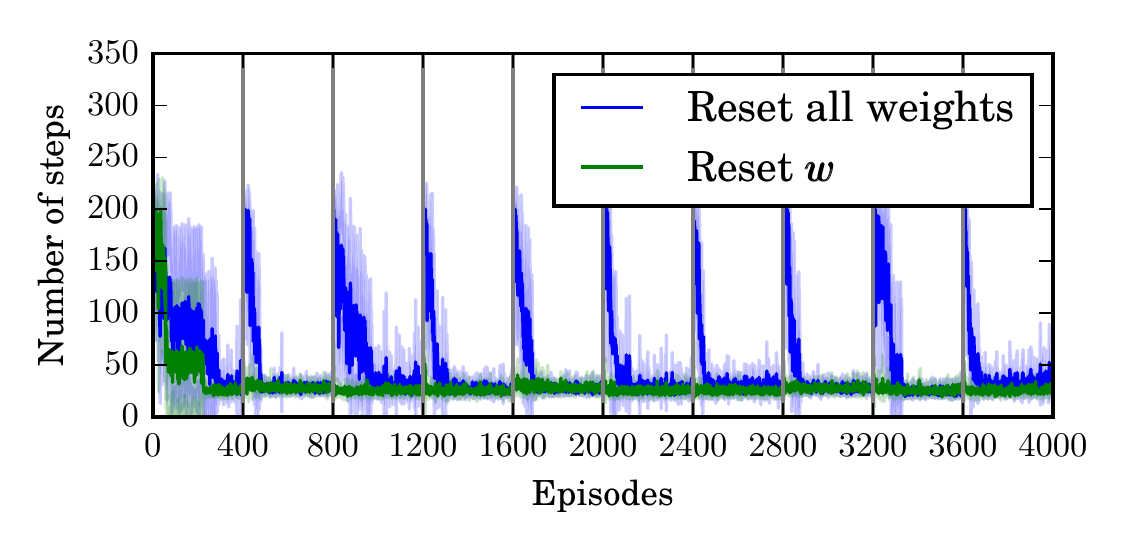}}

\vspace{-.2cm}
\caption{Performance results for repeatedly moving start and goal position by one cell every 400 episodes. A total of three different start and goal positions were used and then repeated.
The episode length was capped at 200 steps.}
\label{fig:tiny-change-comp}
\end{figure*}

\subsection{Multi Task Learning}

The Fitted SF algorithm was also tested in two transfer settings where the start and goal locations are changed periodically between a fixed set of different locations.
Changing the goal location is equivalent to changing the reward function while holding the transition dynamics fixed.

\paragraph{Transfer with Slight Reward Changes}
Figure~\ref{fig:tiny-change-comp-a} compares the episode length of the Fitted Q-iteration implementation and Fitted SF implementation when start and goal locations are moved by one grid cell.
Once the reward function is changed, the $\pmb{w}$ weight parameter of the Fitted SF algorithm is re-initialized to zero.
For Fitted Q-iteration the trained weights are kept after every reward function change.
While initial training is slower for the Fitted SF algorithm, a change in reward function degrades performance significantly less in comparison to Fitted Q-iteration, demonstrating the robustness of the Fitted SF algorithm. 
Figure~\ref{fig:tiny-change-comp-b} compares two different resetting strategies for the Fitted SF learning algorithm: in one run all weights are re-initialized after a reward function change, while in the other the learned SF is kept between training rounds.
One can see that keeping the SF weight matrix $\pmb{\Psi}$ boosts performance significantly.
This verifies the assumption presented by~\citeauthor{barreto2016successor}.

\paragraph{Transfer with Significant Reward Changes} 
To further test if SFs can be used for transfer between different domains, both algorithms are evaluated again on the same grid world, but the goal location is rotated through all four corners of the grid.
The start location is always the corner diagonally across the grid from the goal.
Changing start and goal locations in this way causes the reward function and the optimal policy to change more significantly. 

To further stabilize learning and ensure sufficient exploration, both algorithms select actions using an $\varepsilon$-greedy policy.
The $\varepsilon$ probability is decayed according to the rule $\varepsilon_t = 0.9 \cdot 0.95^t + 0.1$, where $t$ is the episode index.
This episode index $t$ is reset to zero after every reward function change.
Ensuring sufficient exploration allows the Fitted SF algorithm to efficiently re-estimate its reward model.

\begin{figure*}
\centering
\begin{minipage}{.62\textwidth}
\centering
\includegraphics[width=\textwidth]{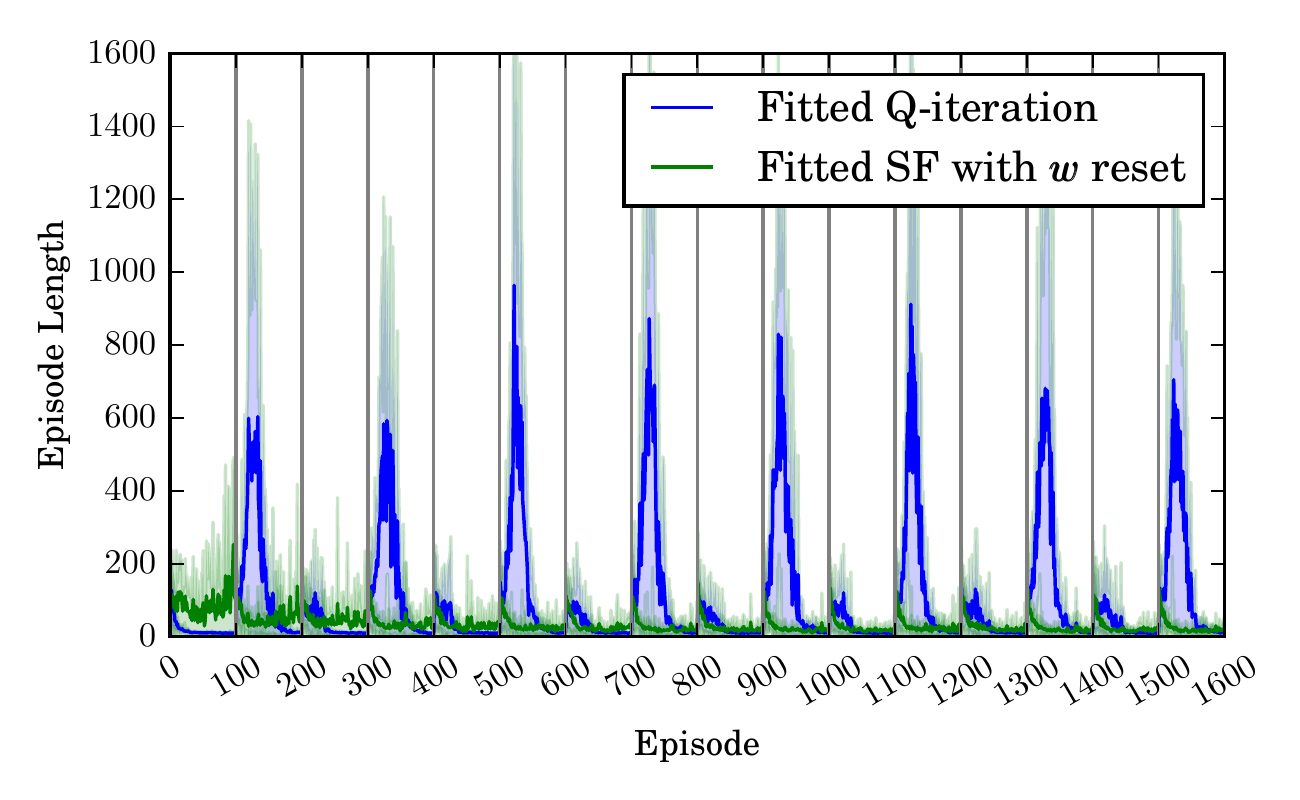}
\vspace{-1.05cm}
\caption{Comparison of the Fitted Q-iteration and Fitted SF algorithm when rotating every 100 episodes the goal location through all four corners of the grid. Fitted Q-iteration uses a learning rate of $0.01$, Fitted SF learning uses a learning rate of $0.01$ for the SF and a learning rate of $0.1$ for the reward model. The episodes were capped at 4000 steps.}
\label{fig:eps-decay}
\end{minipage}~
\begin{minipage}{.38\textwidth}
\begin{tabular}{| p{2.5cm} | l |}
\hline
                             & Avg. Episode Length \\
\hline
 Fitted Q-iteration & $99.46 \pm 10.43$ \\
 \hline
 Fitted SF             & $34.50 \pm 2.17$  \\
 \hline
 \hline
$p$-value & $\mathbf{1.90 \cdot 10^{-17}}$ \\
 \hline
\end{tabular}
\captionof{table}{Average episode length for Figure~\ref{fig:eps-decay}.
The $p$-value of the Welch's t-test tests if the episode lengths are significantly different.}
\label{tab:performance-diff}

\centering
\includegraphics[width=\textwidth]{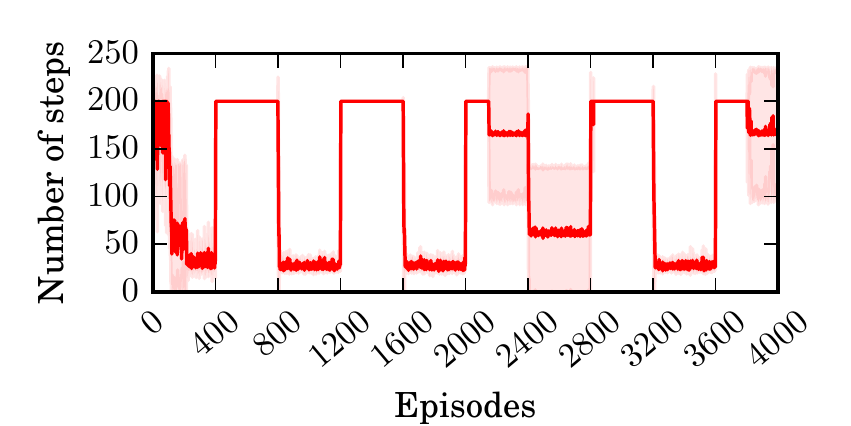}
\vspace{-.7cm}
\captionof{figure}{Episode length of Fitted SF when reward functions change every 400 episodes. The episodes were clipped at 200 steps. All other parameters are the same as in Figure~\ref{fig:eps-decay}}
\label{fig:failure-case}

\end{minipage}
\end{figure*}

\begin{figure}
\centering
\vspace{-0.5cm}
\subfloat[Loss Objective $\mathcal{L}_{SF}$~\eqref{eq:sf-loss}]{\label{fig:eps-decay-lsf}\includegraphics[width=0.24\textwidth]{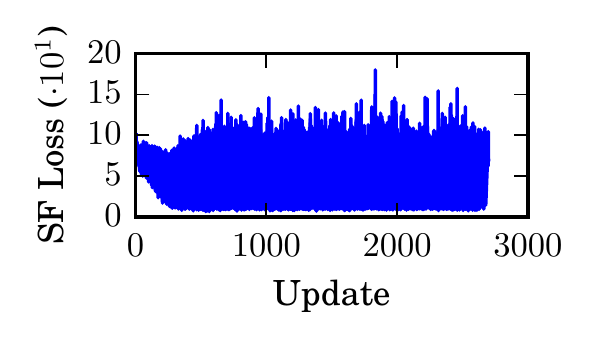}}
\subfloat[Loss Objective $\mathcal{L}_{R}$~\eqref{eq:reward-loss}]{\label{fig:eps-decay-lr}\includegraphics[width=0.24\textwidth]{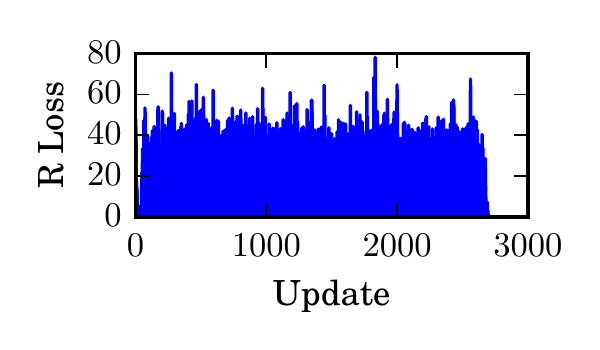}}
\caption{Evolution of the Loss function for the Fitted SF algorithm. A gradient update was applied every 100 steps.}
\label{fig:eps-decay-loss}
\vspace{-0.3cm}
\end{figure}

Figure~\ref{fig:eps-decay} compares the episode length of both algorithms over several repeats of the four goal locations. 
The ordering of the different goal locations is not changed during the experiment. 
One can see that the change in reward function has an impact on both algorithms, but the Fitted SF algorithm outperforms Fitted Q-iteration significantly.
Table~\ref{tab:performance-diff} compares the average episode length across all episodes and shows that our Fitted SF algorithm outperforms the Fitted Q-iteration significantly.
Figure~\ref{fig:eps-decay-loss} shows how the loss functions of the Fitted SF algorithm evolves during the experiment. 
Updates were done only every 100 steps (each gradient update used a batch of 100 transitions).
As expected, the reward loss $\mathcal{L}_R$ does not seem to decrease significantly in a steady way but oscillates instead.
However, the estimates seem to be good enough to achieve a significant performance difference over Fitted Q-iteration.
Interestingly, the SF loss $\mathcal{L}_{SF}$ oscillates during training between very low and high values. 

Figure~\ref{fig:failure-case} shows a failure setting of the Fitted SF algorithm: 
If $\varepsilon = 0.3$ and is not annealed, only the first optimal policy and the first reward function is learned and then preserved across all subsequent changes.
As a result, one can see a learning curve for the first 400 episodes and then Fitted SF hits the episode time-out of 200 steps for the next reward configuration.
If a reward function similar to the first is presented to the agent again, Fitted SF solves this problem easily because it reuses the weights it has learned at the beginning of the experiment.
In other words, Fitted SF is not able to transfer the solution learned in the first 400 episodes to the other tested reward functions.

\section{Discussion}

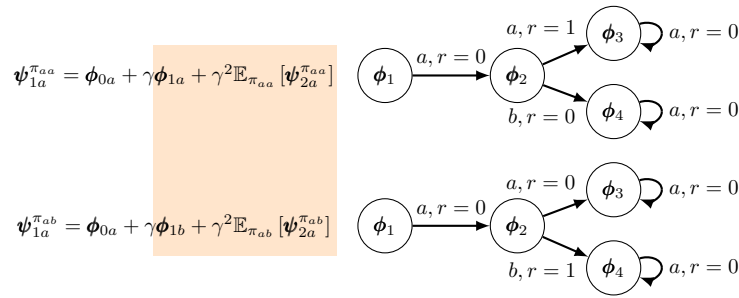
\begin{figure*}
\centering
\input{sf_counter_example.tex}
\caption{Successor Feature Transfer Counter Example. The change in optimal action at state $\pmb{\phi}_2$ causes the SF at state $\pmb{\phi}_1$ to change.}
\label{fig:sf_counter_example}
\end{figure*}

The goal of using SFs is to capture a feature set common to a set of MDPs and this idea seems to perform well for transfer between these MDPs. 
Interestingly, Figure~\ref{fig:eps-decay-lsf} shows that the SF loss objective oscillates despite the fact that the algorithm recovers a near optimal policy quickly.

To get a better understanding why the loss objective oscillates, consider the transfer example shown in Figure~\ref{fig:sf_counter_example}.
In this example, the two MDPs have two actions and deterministic transitions indicated by arrows. 
Rewards are indicated by the arrow labels and the two MDPs only differ in reward for two specific transitions.
This difference in reward causes the optimal policy for each MDP to be different: The policy $\pi_{aa}$, which only selects action $a$, is optimal in the first MDP; the policy $\pi_{ab}$, which selects action $b$ at state $\phi_2$ and action $a$ elsewhere, is optimal in the second MDP.
The left side of Figure~\ref{fig:sf_counter_example} shows the successor feature for both optimal policies, which is different for the two MDPs.
This difference is caused because SFs are constrained to be similar to features the agent sees in the future.
However, which features are seen is governed by the (optimal) policy.
This highlights a key limitation of using Successor Features for transfer: the learned representation is not transferrable between optimal policies.
When solving a previously unseen MDP, a learned SF representation can only be used to initialize the search for an optimal policy and the agent still has to adjust the SF representation to the policy that is only optimal in the current MDP.

The fact that the SF representation has to be re-learned for each individual MDP can be seen in our experiments.
In Figure~\ref{fig:eps-decay-lsf} they contribute to the oscillations of the SF loss objective. 
In the failure case shown in Figure~\ref{fig:failure-case} the SF representation does not transfer at all and instead represents an initialization that the gradient optimizer cannot use to adjust to the new reward function.
This behaviour is not surprising because in this experiment the goal location was changed to a different corner in the grid, causing the optimal policy to change significantly.
In the positive test case shown in Figure~\ref{fig:eps-decay} this is mitigated by resetting the policy first to uniformly random exploration (by annealing $\varepsilon$ from 1.0 to 0.1) which can be thought of as smoothing the transitions between different reward functions.

This result also agrees with the first transfer experiment shown in Figure~\ref{fig:tiny-change-comp}.
Because the reward function and optimal policy is only changed slightly, the SF representations corresponding to each optimal policy and reward function are likely to be very similar.
As a result, the algorithm can adjust to the new reward function very quickly.
\citeauthor{barreto2016successor} also presented empirical results using a variation of Generalized Value Iteration~\cite{Sutton98} on a version of Puddle World~\cite{sutton1996generalization} where the location of the puddle changed slightly.
Their experiment, which shows a significant performance boost by transferring a SF representation, is similar to slight reward change test case because the changes in the reward function did not cause a drastic change in the optimal policy.

\section{Conclusion}

The presented empirical results demonstrate an interesting advantage and dis-advantage of transferring SFs between MDPs that only differ in reward function.
While we were able to show a significant performance boost by using this approach, we also highlighted that the learned feature representation is dependent on the policy they are learned for.
Hence, SF representations are an unsuitable choice in this context because one is typically interested in transferring knowledge between tasks with different optimal policies.

The fact that transferring a SF representation between tasks gives a significant boost in learning speed also suggests that learning a transferrable feature representation might be an interesting direction to pursue.
However, such a feature representation needs to be independent of the task's optimal policy. 

\bibliography{library.bib}
\bibliographystyle{icml2017}

\end{document}

%% file: sf_counter_example.tex
%

\begin{tikzpicture}[domain=-3:3,scale=0.8, every node/.style={transform shape}]

\fill [orange,opacity=.2] (-2.85,2) rectangle (.2,-1.5);

\node[circle,draw=black,minimum size=.9cm](phi1) at (1,    1.5) {$\pmb{\phi}_1$};
\node[circle,draw=black,minimum size=.9cm](phi2) at (3.2, 1.5) {$\pmb{\phi}_2$};
\node[circle,draw=black,minimum size=.9cm](phi3) at (4.8, 2.2) {$\pmb{\phi}_3$};
\node[circle,draw=black,minimum size=.9cm](phi4) at (4.8,.9) {$\pmb{\phi}_4$};

\draw[thick,-latex] (phi1) -- (phi2) node[pos=.5, above] {$a,r=0$};
\draw[thick,-latex] (phi2) -- (phi3) node[pos=.9, above left] {$a,r=1$};
\draw[thick,-latex] (phi2) -- (phi4) node[pos=.9, below left] {$b,r=0$};

\node[anchor=west](phi3out) at (5.6, 2.2) {$a, r=0$};
\path[] (phi3) edge[thick,out=20,in=90] (5.6,2.2)
	   (5.6,2.2) edge[-latex,thick,out=-90,in=-20] (phi3);
	   
\node[anchor=west](phi4out) at (5.6, .9) {$a, r=0$};
\path[] (phi4) edge[thick,out=20,in=90] (5.6,.9)
	   (5.6,.9) edge[-latex,thick,out=-90,in=-20] (phi4);

\node[](sf) at (-2.5,    1.5) {$\pmb{\psi}_{1a}^{\pi_{aa}} = \pmb{\phi}_{0a} + \gamma \pmb{\phi}_{1a} + \gamma^2 \mathbb{E}_{\pi_{aa}} \left[ \pmb{\psi}_{2a}^{\pi_{aa}} \right]$};

\node[circle,draw=black,minimum size=.9cm](phi1) at (1,    -1.) {$\pmb{\phi}_1$};
\node[circle,draw=black,minimum size=.9cm](phi2) at (3.2, -1.) {$\pmb{\phi}_2$};
\node[circle,draw=black,minimum size=.9cm](phi3) at (4.8, -.4) {$\pmb{\phi}_3$};
\node[circle,draw=black,minimum size=.9cm](phi4) at (4.8,-1.7) {$\pmb{\phi}_4$};

\draw[thick,-latex] (phi1) -- (phi2) node[pos=.5, above] {$a,r=0$};
\draw[thick,-latex] (phi2) -- (phi3) node[pos=.9, above left] {$a,r=0$};
\draw[thick,-latex] (phi2) -- (phi4) node[pos=.9, below left] {$b,r=1$};

\node[anchor=west](phi3out) at (5.6, -.4) {$a, r=0$};
\path[] (phi3) edge[thick,out=20,in=90] (5.6,-.4)
	   (5.6,-.4) edge[-latex,thick,out=-90,in=-20] (phi3);
	   
\node[anchor=west](phi4out) at (5.6, -1.7) {$a, r=0$};
\path[] (phi4) edge[thick,out=20,in=90] (5.6,-1.7)
	   (5.6,-1.7) edge[-latex,thick,out=-90,in=-20] (phi4);

\node[](sf) at (-2.5,    -1.) {$\pmb{\psi}_{1a}^{\pi_{ab}} = \pmb{\phi}_{0a} + \gamma \pmb{\phi}_{1b} + \gamma^2 \mathbb{E}_{\pi_{ab}} \left[ \pmb{\psi}_{2a}^{\pi_{ab}} \right]$};

\end{tikzpicture} 